\documentclass[runningheads]{llncs}
\usepackage{graphicx}

\usepackage{tikz}
\usepackage{comment} 
\usepackage{amsmath,amssymb} 
\usepackage{color}

\usepackage{import}
\usepackage{gensymb}
\usepackage{booktabs}
\usepackage{multirow}
\usepackage{xcolor}
\usepackage{array}

\usepackage{soul}
\usepackage{microtype}
\usepackage[font=small,labelfont=bf]{caption}
\usepackage{booktabs}
\usepackage{tabularx}
\usepackage{sidecap}

\usepackage{xr}

\newcommand{\image}[1]{
\begin{minipage}[b][][b]{.26\textwidth}
\includegraphics[width=\linewidth]{#1}
\end{minipage}
}

\newcommand{\imageF}[1]{
\begin{minipage}[b][][b]{.23\textwidth}
\includegraphics[width=\linewidth]{#1}
\end{minipage}
}

\newcommand{\imageT}[1]{
\begin{minipage}[b][][b]{.49\textwidth}
\includegraphics[width=\linewidth]{#1}
\end{minipage}
}

\definecolor{mygray}{gray}{0.6}

\newcommand{\gray}[1]{\textcolor{mygray}{#1}}
\definecolor{myolive}{RGB}{85,107,47}

\newcommand{\fig}[1]{Fig. \ref{fig:#1}}
\newcommand{\eqn}[1]{Eqn. \ref{eqn:#1}}
\newcommand{\tab}[1]{Table \ref{tab:#1}}
\makeatletter
\DeclareRobustCommand\onedot{\futurelet\@let@token\@onedot}
\def\@onedot{\ifx\@let@token.\else.\null\fi\xspace}

\makeatother

\usepackage{mathtools}

\newcommand{\etal}{\textit{et al}. }
\newcommand{\ie}{\textit{i}.\textit{e}. }
\newcommand{\eg}{\textit{e}.\textit{g}. }
\newcommand{\wrt}{w.r.t. }

\newcommand{\etc}{\emph{etc}. }

\newcommand{\norm}[1]{\left\lVert#1\right\rVert}

\makeatletter
\newcommand*{\inlineequation}[2][]{%
  \begingroup
    \refstepcounter{equation}%
    \ifx\\#1\\%
    \else
      \label{#1}%
    \fi
    \relpenalty=10000 %
    \binoppenalty=10000 %
    \ensuremath{%
      #2%
    }%
    ~\@eqnnum
  \endgroup
}
\makeatother

\usepackage{hyperref}

\begin{document}
\pagestyle{headings}
\mainmatter
\def\ECCVSubNumber{1337}  

\title{Single View Metrology in the Wild} 

\titlerunning{Single View Metrology in the Wild}
%
\author{Rui Zhu\inst{1}
Xingyi Yang\inst{1} \and
Yannick Hold-Geoffroy\inst{2} \and
Federico Perazzi\inst{2} \and
Jonathan Eisenmann\inst{2} \and
Kalyan Sunkavalli\inst{2} \and
Manmohan Chandraker\inst{1}}
\authorrunning{R. Zhu, X. Yang et al.}
%
\institute{University of California San Diego, La Jolla CA 92093, USA\\
\email{\{rzhu,x3yang,mkchandraker\}@eng.ucsd.edu}\and
Adobe Research, San Jose CA 95110, USA\\
\email{\{holdgeof,perazzi,eisenman,sunkaval\}@adobe.com}}
\maketitle

\begin{figure}
\centering
\includegraphics[width = 1.\linewidth]{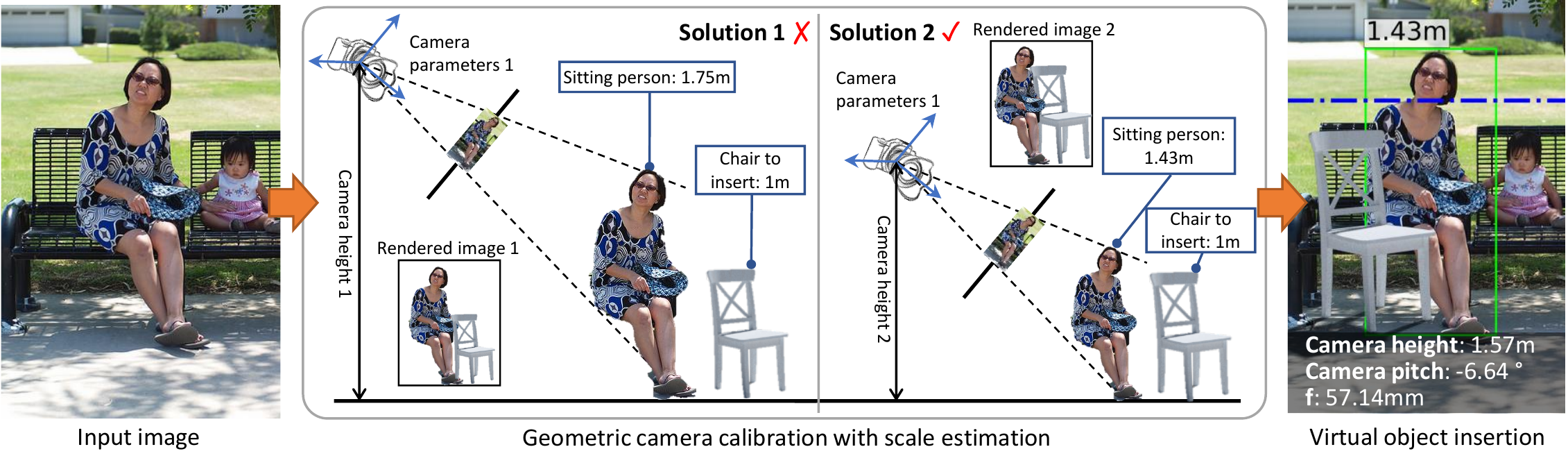}
\caption{Given the image on the left, single view metrology can recover the scene and the camera parameters in 3D only up to a global scale factor (for example, the two solutions in the middle). Our method accurately estimates absolute 3D camera parameters and object heights (middle, left) to produce realistic object insertion results (right).}
\vspace{-0.4cm}
\label{fig:scene_scale_illu}
\end{figure}

\begin{abstract}



Most 3D reconstruction methods may only recover scene properties up to a global scale ambiguity.
We present a novel approach to single view metrology that can recover the \emph{absolute} scale of a scene represented by 3D heights of objects or camera height above the ground as well as camera parameters of orientation and field of view, using just a monocular image acquired in unconstrained condition. Our method relies on data-driven priors learned by a deep network specifically designed to imbibe weakly supervised constraints from the interplay of the unknown camera with 3D entities such as object heights, through estimation of bounding box projections. We leverage categorical priors for objects such as humans or cars that commonly occur in natural images, as references for scale estimation. We demonstrate state-of-the-art qualitative and quantitative results on several datasets as well as applications including virtual object insertion. Furthermore, the perceptual quality of our outputs is validated by a user study.\footnote{Parts of this work were done when Rui Zhu was an intern at Adobe Research.}

\keywords{Single view metrology, absolute scale estimation, camera calibration, virtual object insertion}
\end{abstract}
\section{Introduction} \label{sec:intro}
Reconstructing a 3D scene from images is a fundamental problem in computer vision.
Despite many successes on this task, most previous works only reconstruct scenes up to an unknown scale.
This is true for many problems including structure-from-motion (SfM) from uncalibrated cameras~\cite{HartleyZissermanMVG}, monocular camera calibration in the wild \cite{workman2015deepfocal,yannick2018perceptual,workman2016horizon} and single image depth estimation~\cite{eigen2014depth,li2018megadepth}. This ambiguity is inherent to the projective nature of image formation and resolving it requires additional information. For example, the seminal work ``Single View Metrology'' of Criminisi et al.~\cite{criminisi2000single} relies on the size of reference objects in the scene.

In this work, we consider the problem of single view metrology ``in the wild'', where only a single image is available for an unconstrained scene composed of objects with unknown sizes.
In particular, we plan to achieve this via geometric camera calibration with absolute scale estimation, \ie recovering camera orientation (alternatively, the horizon in the image),  field-of-view, and the \emph{absolute} 3D height of the camera from the ground.
Given these parameters, it is possible to convert any 2D measurement in image space to 3D measurements.

Our goal is to leverage modern deep networks to build a robust, automatic single view metrology method that is applicable to a broad variety of images.
One approach to this problem could be to train a deep neural network to predict the scale of a scene using a database of images with known absolute 3D camera parameters. 
Unfortunately, no such large-scale dataset currently exists. 
Instead, our insight is to leverage large-scale datasets with 2D object annotations ~\cite{mscoco,xiang2014pascal,geiger2013kitti,eigen2014depth}. 
In particular, we make the observation that objects of certain categories such as humans and cars are ubiquitous in images in the wild~\cite{mscoco,xiang2014pascal} and would make good ``reference objects'' to infer the 3D scale.

While the idea of using objects of known classes as references to reconstruct camera and scene 3D properties has been used in previous work~\cite{hoiem2008putting,kar2015amodal}, we significantly extend this work by making fewer approximations in our image formation model (\eg full perspective camera vs. zero camera pitch angle, infinite focal length in~\cite{hoiem2008putting}), leading to better modeling of images in the wild.
Moreover, our method learns to predict all camera and scene properties (object and camera height estimation, camera calibration) in an end-to-end fashion; in contrast, previous work relies on individual components that address each sub-task.
We demonstrate that this holistic approach leads to state-of-the-art results across all these tasks on a variety of datasets (SUN360, KITTI, IMDB-23K).
We also demonstrate the use of our method for applications such as virtual object insertion, where we may automatically create semantically meaningful renderings of a 3D object with known dimensions (see Fig.~\ref{fig:scene_scale_illu}).

In summary, we propose the following contributions:
\begin{itemize}
    \item A state-of-the-art Single View Metrology method for images in the wild that performs  geometric camera calibration with absolute scale---horizon, field-of-view, and 3D camera height---from a monocular image.  
    \item A weakly supervised approach to train the above method with only 2D bounding box annotations by using an in-network image formation model. 
    \item Application to scale-consistent object insertion in unconstrained images.
\end{itemize}

\section{Related Work}

\subsubsection{Camera calibration} To estimate the camera parameters, numerous efforts have been made for estimating camera intrinsics~\cite{workman2015deepfocal,chen2004cameracalib,deutscher2002automatic,yannick2018perceptual} by explicit reasoning or learning in a data-driven fashion. In addition, to estimate camera extrinsics, \eg camera rotation angles or in the form of horizon estimation, classical methods~\cite{zhai2016detecting,barinova2010geometric,denis2008efficient,lee2013automatic} look for low-level features such as line segments. More recently, methods are proposed to directly regress the horizon from the input image~\cite{man2018groundnet,workman2016horizon,kluger2020horizon} by learning from large-scale datasets annotated with ground truth horizons. The human sensitivity to calibration errors is studied in~\cite{yannick2018perceptual}.


\subsubsection{Depth prediction in the wild} As we discussed in Section \ref{sec:intro}, the problem of scene scale estimation will be solved if we are able to predict pixel-wise depth for the scene. There has been a line of work in this topic. For domains which we can acquire the ground truth absolute depth with depth sensors we may learn to predict depth in a supervised fashion~\cite{eigen2014depth,wang2018learning,kim2019pedx}. However given the limitation of the range or mobility of the sensors, these datasets are more or less limited to specific scenes. In other cases, people are able to acquire ground truth from stereo matching~\cite{garg2019learning,geiger2013vision,wang2018deeplens} but a large-scale stereo depth dataset for images in the wild is still absent. Other people have turn to proxy methods for collecting depth via structure-from-motion (SfM)~\cite{li2018megadepth,li2019learningdepthman}, or in the form of relative depth~\cite{chen2016depthwild}, or from synthetic images~\cite{martinez2018beyond,atapour2018real}. However, these methods either produce depth without absolute scale, or pose a domain gap to natural images.

\subsubsection{Single view metrology} Another line of work that seeks to estimate 3D scene parameters from images is Single View Metrology~\cite{criminisi2000single}, which recovers scene structure in 3D from purely 2D measurements. These methods look for 2D properties such as vanishing lines and vanishing points as well as object locations, to establish relations among 3D sizes of objects in the image based on 2D measurements. Some works have been done to embed Single View Metrology in a framework to estimate the size of an unknown object in the scene or the camera height itself~\cite{andalo2015efficient,kar2015amodal,hoiem2008putting,lalonde2007photo}, given at least one reference object with known size.


\section{Method}

\begin{figure}[!!t]
\centering
\includegraphics[width = 0.6\linewidth]{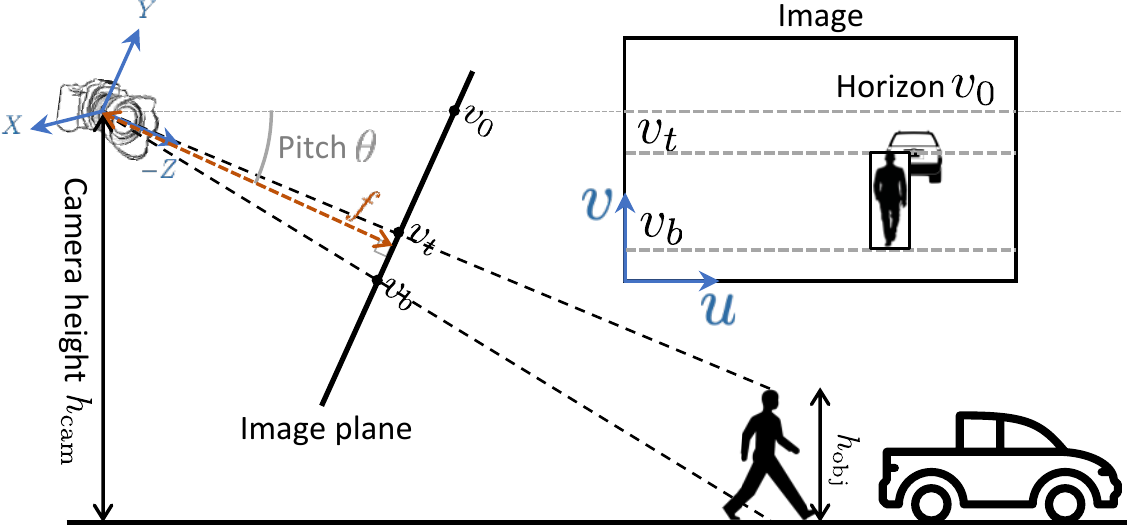}
\caption{Camera model of the scene (bottom) and measurements in image space (top).}
\label{fig:scene_model}
\end{figure}

\subsection{Recovering 3D Parameters from 2D Annotations}

We start by describing the image formation model that allows us to associate 3D camera parameters, 3D object sizes (\ie heights) and 2D bounding boxes. This is also illustrated in Fig.~\ref{fig:scene_model}.

We assume the world is composed of a dominant ground plane on which all objects are situated, and a camera that observes the scene. 
We adopt a perspective camera model similar to~\cite{hoiem2008putting,yannick2018perceptual}, which is parameterized by camera angles (yaw $\varphi$, pitch $\theta$ and roll $\psi$), focal length $f$ and camera height $h_{\mathrm{cam}}$ to the ground (see \fig{scene_model}). 
For the measurements in the vertical axis of image frame, the location of the horizon is $v_0$, while the vertical image center is at $v_c$. 
Each object bounding box have a top $v_t$ and bottom $v_b$ location in the image. 
We assume all images were taken with zero roll, or were rectified beforehand~\cite{lee2013automatic}. We further assume, without loss of generality, a null yaw and zero distortion from rectification. Camera pitch $\theta$ can be computed from $v_c$, $v_0$ and $f$ using 
\begin{align}
    \theta &= \label{eqn:pitch}
    \arctan{\frac{v_c - v_0}{f}}.
\end{align}






Consider a thin object of height $h_{\mathrm{obj}}$ with its bottom located at $[x, 0, z]^T$ in world coordinates and its top at $[x, h_{\mathrm{obj}}, z]^T$. These points project to $[u,  v_b]^T$ and $[u, v_t]^T$ respectively in the camera coordinates. 
Based on the perspective camera model (see Eqn. 3 of the supplementary material), we have
\begin{align}
    v_t &= \label{eqn:vt}
    \frac{
        (f\cos\theta + v_c\sin\theta) h_{\mathrm{obj}}
         + (-f\sin\theta + v_c\cos\theta)z - fh_{\mathrm{cam}}
        }
    {h_{\mathrm{obj}}\tan\theta + z\cos\theta}, \\
    v_b &= \label{eqn:vb}
    \frac{-f\sin\theta z + v_c\cos\theta z - f h_{\mathrm{cam}}}
    {z\cos\theta},
\end{align}

\noindent where $[u_c, v_c] \in \mathbb{R}^2$ is the camera optical center which we assume is known. Substituting $z$ from \eqn{vb} into \eqn{vt}, we may derive $v_t$ from camera focal length $f$, pitch $\theta$, camera height $h_{\mathrm{cam}}$ and object height $h_{\mathrm{obj}}$. Hoeim et al. \cite{hoiem2008putting} make a number of  approximations including $\cos\theta\approx1$, $\sin\theta\approx \theta$ and $(v_c-v_0) \times (v_c - v_0)\//{f^2} \approx 0$, to linearly solve for the object height:

\begin{align} \label{eqn:yh}
    h_{\mathrm{obj}} &= h_{\mathrm{cam}} \frac{v_t - v_b}{v_0 - v_b}.
\end{align}
In contrast, we model the full expression accounting for all the camera parameters.

\eqn{vt} and \eqn{vb} establish a relationship between the camera parameters, including its 3D height, the 3D heights of objects in the scene, and the 2D projections of these objects into the image. Moreover, note that once we can estimate these parameters, we can directly infer the 3D size of \emph{any} object from its 2D bounding box, thus resolving the scale ambiguity in monocular reconstruction. In the following section, we introduce ScaleNet, a deep network that leverages this constraint to create weak supervision to learn to predict 3D camera height.

\subsection{ScaleNet: Single View Metrology Network with Absolute Scale Estimation}
\label{sec:pipeline}

\begin{figure}[!!t]
\centering
\includegraphics[width = 1.\linewidth]{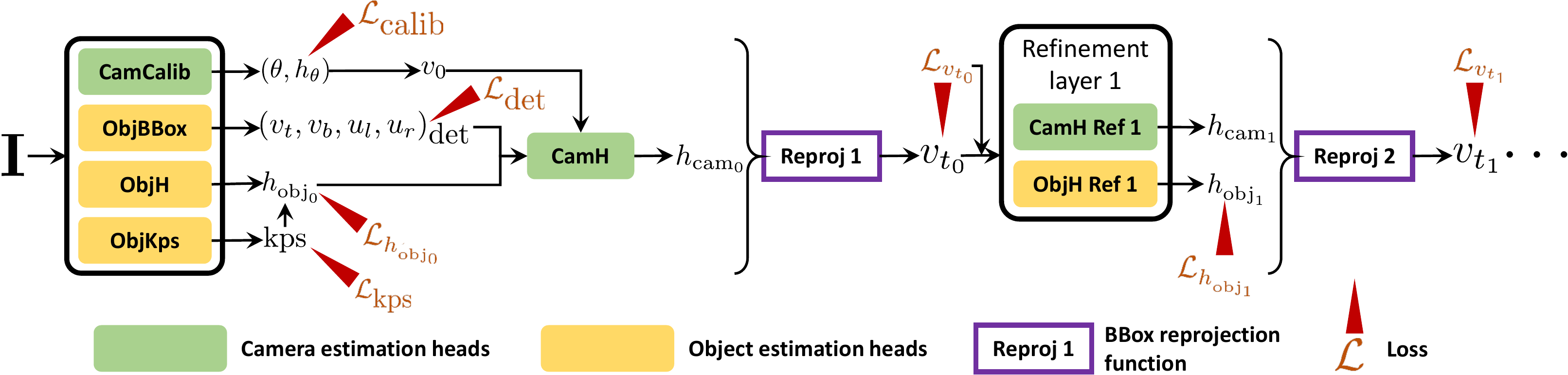}
\caption{Overview of our method. From the input image $\textbf{I}$, the camera calibration module estimates pitch $\theta$ and field of view $h_\theta$, and the object estimation modules estimate keypoints for person, object heights $h_\textrm{obj}$ and bounding boxes. The estimated horizon $v_0$, bounding boxes and object heights are fed into the camera height estimation module to give an initial estimation $h_\textrm{cam}$. The bounding box reprojection errors $\mathcal{L}_{v_t}$ are then computed from the reprojection module (see \eqn{vt} and \eqn{l1_loss}), and together with other variables are fed to the refinement network to estimate updates on the camera height and object heights. Several layers of refinement are made to produce the final estimation.}
\label{fig:arch}
\end{figure}

Previous work~\cite{hoiem2008putting,kar2015amodal} has shown that when scene parameters (\eg camera parameters, object sizes) are reasonable, reprojected 2D bounding boxes should ideally fit the detected ones in the image frame. 
We follow a similar path in our weakly-supervised learning framework and specifically focus on humans and cars, given that they are the most commonly occurring object categories in datasets of images in the wild (\eg COCO dataset~\cite{mscoco}). 

Our end-to-end method, referred to as \textbf{ScaleNet (SN)}, is split into two parts, which we describe in \fig{arch}. First, all the object bounding boxes and camera parameters except camera height are jointly estimated by a geometric camera calibration network. 
These parameters are directly supervised during training. 
Second, a cascade of PointNet-like networks~\cite{ranftl2018deepf} estimates and refines the camera height (scene scale) based on the previous outputs. This second part is weakly supervised at each stage using a bounding box reprojection loss.

\subsubsection{Camera calibration module and object heads}
\label{sec:calib_module}

The camera calibration module is inspired by~\cite{yannick2018perceptual}, where we replace their backbone with Mask R-CNN~\cite{he2017maskpaper,massa2018mrcnncode}, to which we add heads to estimate the camera parameters. To train the camera calibration module, we follow the representation of~\cite{yannick2018perceptual}, including bins and training loss. However, instead of predicting the focal length $f$ (in pixels), we find it easier to predict the vertical field of view $h_\theta$, which can be converted to the focal length and the horizon midpoint using:
%
\begin{align*}
    f &= \frac{ \frac{1}{2}h_{\textrm{im}} }{ \tan\left({\frac{1}{2}h_\theta}\right) }, \quad v_0 = \frac{ \frac{1}{2}h_{\textrm{im}}\tan\theta } { \tan\left({\frac{1}{2}h_\theta}\right) } + \frac{h_{\textrm{im}}}{2},
\end{align*}
where $h_{\textrm{im}}$ is the height of the image in pixels. We also use additional heads to estimate the object bounding box, height and person keypoints (since we find that a person's 3D height in an image is closely related to their pose) from ROI features which share the same backbone as the camera estimation module. Please refer to the supplementary material for the full architecture of this network. In total, we enforce three losses on this part of the model, \ie the camera calibration loss $\mathcal{L}_{\textrm{calib}}\big(\theta, h_\theta \big)$ and the detection losses $\mathcal{L}_{\textrm{det}}$ and $\mathcal{L}_{\textrm{kps}}$.

The $i_{th}$ object with detected 2D top position $v_{t_\textrm{det}}^i$ is reprojected to the image by \eqn{vt} to $v_t^i$ with estimated object height, camera height and camera parameters, and we define the bounding reprojection error as

\begin{align} \label{eqn:l1_loss}
    &\mathcal{L}_{v_t}\Big({\{v_t^i\}_{i=1}^N}\Big) = \frac{1}{N} \sum_{i=1}^N \norm{v_{t_\textrm{det}}^i - v_t^i}.
\end{align}

\subsubsection{Object height prior}
\label{sec:height_prior}

The above bounding box supervision has the same scale ambiguity as previous work.
However, explicitly modeling 3D object heights allows us to use a prior on size to regularize the network to produce a meaningful object height estimation. 
We follow~\cite{hoiem2008putting} and use a Gaussian prior fit from statistics (\ie for 1.70$\pm$0.09m for people and 1.59$\pm$0.21m for cars). For an object $i$ of height $h_{\textrm{obj}}^i$ and prior Gaussian distribution $\mathcal{P}(x; \mu, \sigma)$, we define the height prior loss as

\begin{align} \label{eqn:prior}
    \mathcal{L}_{h_{\textrm{obj}}}\Big({\{h_{\textrm{obj}}^i\}_{i=1}^N}\Big)= - \frac{1}{N} \sum_{i=1}^N \mathcal{P}(h_{\textrm{obj}}^i; \mu, \sigma).
\end{align}

\subsubsection{Camera height estimation module}
\begin{figure}[!!t]
\centering
\includegraphics[width = 0.7\linewidth]{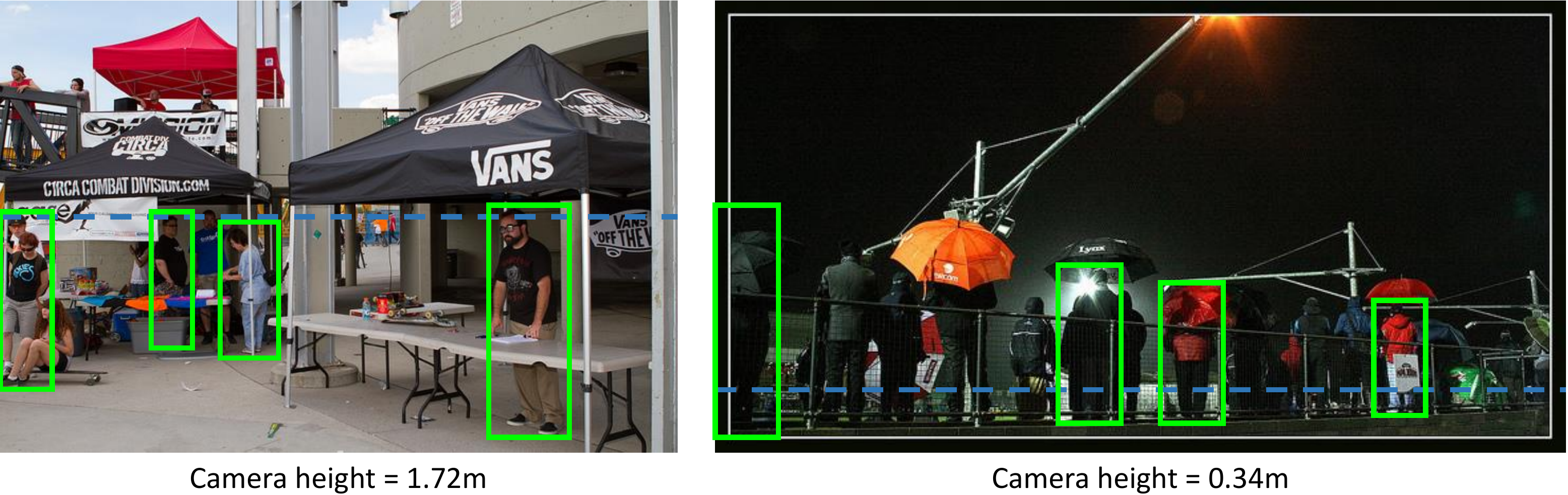}
\caption{Example of images with different camera heights exhibiting similar bounding boxes.}
\label{fig:camH_pred_example}
\end{figure}

Directly predicting camera height from images would require the network to learn to be robust to a wide variety of appearance properties (object, layout, illumination, \etc). Instead, we design a camera height estimation module that leverages the strong \emph{geometric} relationship between camera height, 2D bounding boxes and the horizon. As exemplified in \fig{camH_pred_example}, both images are composed of a group of standing people while the horizons are not fully visible in the back. At first glance, both images seem to have the same camera orientation, since the people take roughly the same space in the image. However the camera height is quite different between both images. Based on this observation, instead of estimating camera height from image appearance, we take advantage of middle level representations of the scene (\eg object bounding boxes and estimated horizon line) and feed those to the camera height estimation module which is derived from PointNet~\cite{qi2017pointnet}. Its input is the concatenation of all object bounding box coordinates and the offset between the bounding box and horizon, \ie $\gamma_0 = [v_0, u_{l_\textrm{det}}, u_{r_\textrm{det}}, v_{t_\textrm{det}}, v_{b_\textrm{det}}, v_{t_\textrm{det}}-v_0, v_{b_\textrm{det}}-v_0, h_{\textrm{obj}}]^T  \in \mathbb{R}^8$ where $u_{l_\textrm{det}}$ and $u_{r_\textrm{det}}$ are the left and right coordinates of the detected bounding box. The network outputs the camera height as a discrete probability distribution. Finally, a weighted sum after a \texttt{softmax} is applied to obtain the camera height estimation.

\subsubsection{Cascade refinement layers}
\begin{figure}[!!t]
\centering
\includegraphics[width = 0.7\linewidth]{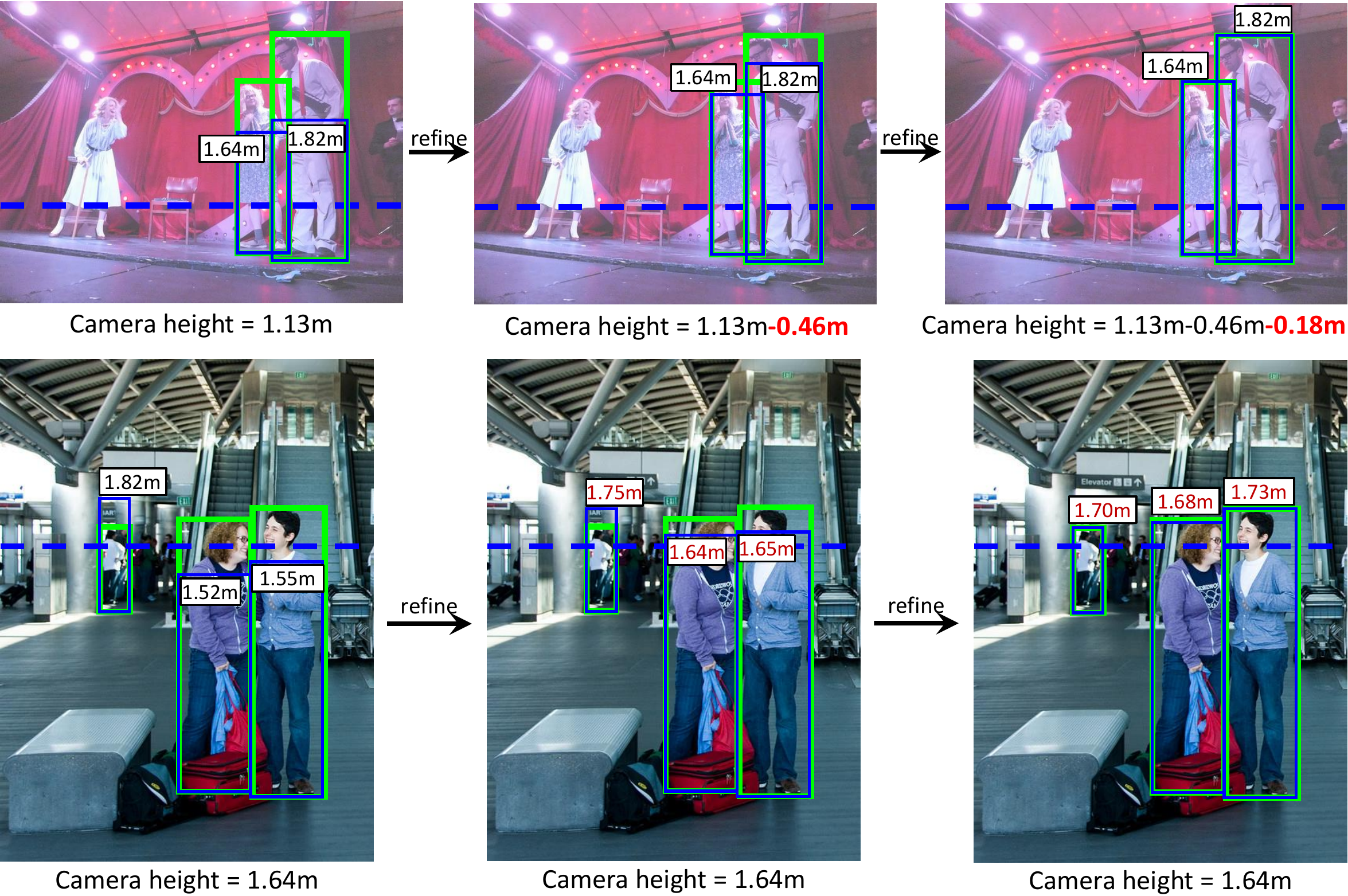}
\caption{Example of cascade refinements of camera height (top) and person heights (bottom). The refined parameters are labelled in red.}
\label{fig:cascade_example}
\end{figure}

We observe that we can iteratively refine the camera height by considering all scene parameters jointly. Inspired by the cascade refinement scheme from~\cite{ranftl2018deepf}, we propose to look at the error residual---in our case the object bounding box reprojection error---and predict a \emph{difference} to the estimated parameters. The whole process is highlighted in \fig{cascade_example}, where the reprojected object bounding boxes are shorter than the detected ones at first. After a first step of refinement, the network reduces the camera height to reduce the object bounding boxes error, and so on. To this end, we design layers of refinement, where in layer $j \in \{{1, 2, \dots, M}\}$ a camera height and object height refinement module takes as input the object bounding box reprojection residuals and the other camera parameters, formally as $\gamma_j = [v_0, (u_l^i, u_r^i, v_{t_{j-1}}^i, v_b^i)_\textrm{det}, v_{t_{j-1}}^i-v_{t_\textrm{det}}^i, h_{\textrm{obj}_{j-1}}^i, h_{\textrm{cam}_{j-1}}]^T  \in \mathbb{R}^8$ where $i \in \{1, 2, \dots N\}$ is the object index. Each refinement layer $j$ predicts updates $\Delta h_{\textrm{cam}_j}$ and $\Delta h_{\textrm{obj}_j}^i$ so that $h_{\textrm{obj}_j}^i = \Delta h_{\textrm{obj}_j}^i + h_{\textrm{obj}_{j-1}}^i$ and $h_{\textrm{cam}_j} = \Delta h_{\textrm{cam}_j} + h_{\textrm{cam}_{j-1}}$. An object bounding box reprojection loss $\mathcal{L}_{v_{t_j}}$ and object height prior $\mathcal{L}_{h_{\textrm{obj}_j}}$ are enforced for each layer.

The final training loss is a weighted combination of the losses, written as
\begin{align} \label{eqn:training_loss} 
    \resizebox{0.93\hsize}{!}{
    $\mathcal{L}\left({\big\{{\{v_{t_j}^i\}_{i=1}^N, \{{h_{\textrm{obj}_j}}^i\}_{i=1}^N}\big\}}_{j=0}^M, \theta, f, h_\theta, (...)_{\textrm{det}}, \textrm{kps}\right) = 
      \alpha_1 \sum_{j=1}^M  \mathcal{L}_{v_{t_j}}
    + \alpha_2 \sum_{j=0}^M \mathcal{L}_{{h_{\textrm{obj}_j}}}
    + \alpha_3 \mathcal{L}_{\textrm{calib}}
    + \alpha_4 \mathcal{L}_{\textrm{det}}
    + \alpha_5 \mathcal{L}_{\textrm{kps}}$,
    }
\end{align}
\noindent where $\alpha_{1..5}$ are weighting constants to balance the losses during training.

\subsection{Datasets}

In the following, we describe the datasets and their preprocessing used for training and evaluation. Data generation details, statistics, sampled visualization of all datasets can be found in the supplemental material.

\subsubsection{Calib: Camera calibration dataset}

To train the camera calibration module to estimate camera pitch and field of view from a single image, we follow the data generation pipeline from~\cite{yannick2018perceptual}, where data are cropped from the SUN360 database~\cite{sun360} of 360\degree\ panoramas with sampled camera parameters. We split the resulting camera calibration dataset into 397,987 images for training and 2,000 images for validation. For simplicity, we refer to this dataset as the \texttt{Calib} dataset.

\subsubsection{COCO-Scale: Scale estimation dataset from COCO}

While the \texttt{Calib} dataset provides a large and diversified dataset of images and many ground truth camera pitch and field of view parameters, it does not provide camera height. To complement this dataset, we use the COCO dataset~\cite{mscoco}, which allows us to train our method in a weakly-supervised way. This dataset features 2D annotations of object bounding boxes, keypoints of person, and stuff annotation. We further extend these annotations by using Mask R-CNN~\cite{he2017maskpaper} to infer objects of certain categories, \eg person and car. These additional annotations complement the ones provided in the dataset, which together form our candidate object set. We refer to this dataset as the \texttt{COCO-Scale} dataset.

We filter out invalid objects which do not satisfy our scene model, using the stuff annotation (\eg ground, grass, water) to infer the support relationship of an object with its surrounding pixels; we only keep objects that are most likely situated on a plane (\eg people standing on grassland, cars on a street). For the \textit{person} category, we use the detected keypoints from Mask R-CNN~\cite{massa2018mrcnncode} to detect people with both head and ankles visible to ensure the obtained bounding box is amodal as in \cite{kar2015amodal}. We further filter the images based on aspect ratio, object size and number of objects to keep bounding boxes of certain shape.

This pruning step yields 10,547 training images, 2,648 validation images, taken from COCO's \texttt{train2017} and \texttt{val2017} splits respectively. We further obtain test test images from \texttt{val2017} and ensure no overlap exists between the splits. We call this person-only subset \texttt{COCO-Scale-person}.
For the multi-category setting (\texttt{COCO-Scale-mulCat}), we look for images including both cars and people, which provides us with 12,794 images for training, 3,189 for validation, and 584 for testing.



\subsubsection{KITTI}

We use the KITTI~\cite{geiger2013kitti} dataset to evaluate our camera and object height estimations. We apply the same filtering rules as used on COCO, yielding 298 images for person-only setting (\texttt{KITTI-person}), and 234 images for the multi-category setting (\texttt{KITTI-mulCat}).

\subsubsection{IMDB-23K celebrity dataset for person height evaluation}

\begin{figure}[!!t]
\centering
\includegraphics[width = 0.7\linewidth]{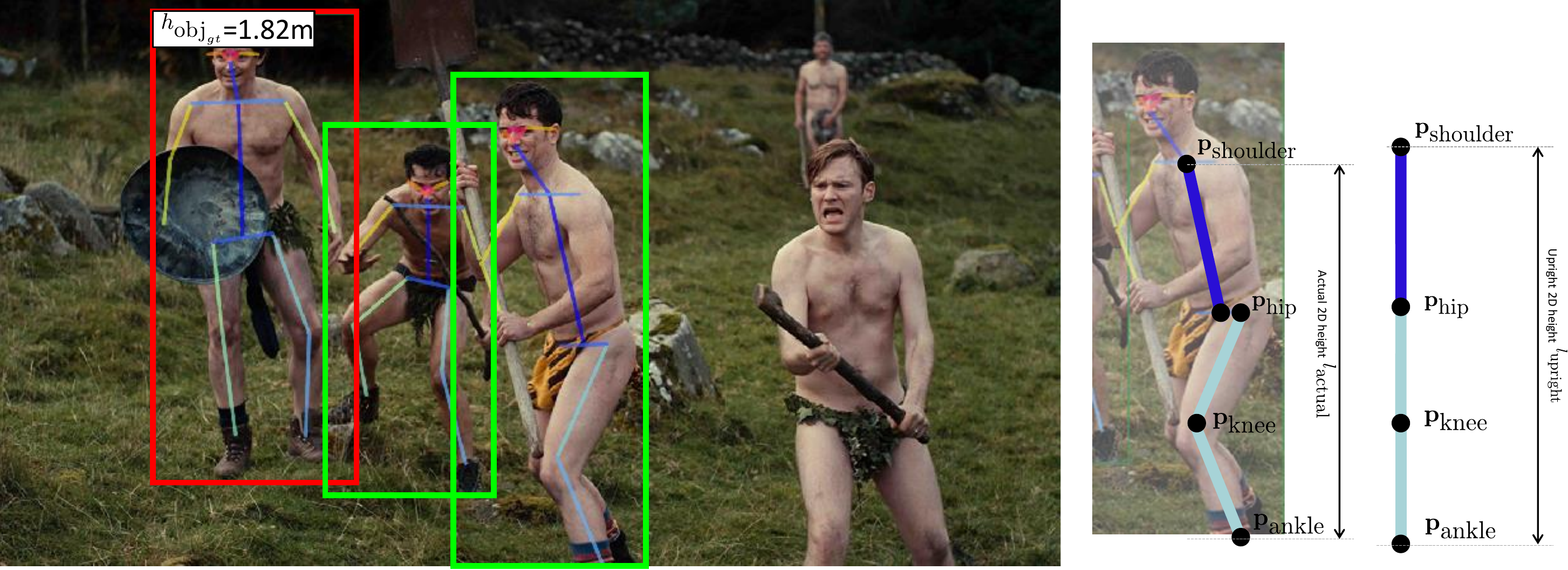}
\caption{(Left) Annotated person bounding box (red) with ground truth height and detected person (green) with keypoints (colored). (Right) Calculation of upright ratio.}
\label{fig:ratio_illu}
\end{figure}

IMDB-23K~\cite{gunel2018celebrity} is a collection of online images of celebrities, with annotations of body height from the IMDB website. We use this dataset to evaluate our object height prediction. However, these height annotations may not be exact and we treat them as \textit{pseudo} ground truth to draw comparisons. We apply the same filtering rules as on COCO, and the filtered dataset consists of 2,550 test images with one celebrity labelled with height in each image. An image from this dataset is shown in \fig{ratio_illu}.

\section{Experiments}

\subsection{Baseline methods}


We use Hoiem \etal~\cite{hoiem2008putting} as the baseline method. For fair comparison, we employ our object proposals or top predictions from Mask R-CNN~\cite{massa2018mrcnncode} as input to this method to replace the original detector~\cite{murphy2003graphical}, which enhances the original method. We set up 2 baseline models: (1) \emph{PGM}: the original model based on a Probabilistic Graphical Model, which takes in object proposals and surface geometry, and predicts camera height and horizon. Object heights can be computed from \eqn{yh} by directly minimizing the reprojection error; (2) \emph{PGM-fixedH}: same as PGM but assumes all objects have canonical height. For people, we use 1.7m which is the mean of the person height prior used in~\cite{hoiem2008putting}). For cars we use 1.59m.

\subsection{Training}

We train our model in two stages. Firstly, we train the camera calibration network with full supervision from the Calib dataset using camera calibration losses and the detection \& keypoint estimation heads with full supervision from COCO ground truth with losses  following~\cite{massa2018mrcnncode}. The backbone, camera calibration head, and detection \& keypoint heads are initialized with a pre-trained Mask R-CNN model~\cite{massa2018mrcnncode}. 

For the second stage, the object height estimation module is trained together while other modules are finetuned, in a weakly-supervised fashion, with the full loss in \eqn{training_loss}. Training details can be found in the supplemental material.

\subsubsection{Variants of ScaleNet (SN)}
We evaluate several variations of ScaleNet~(\emph{SN}): (1) \emph{SN-L1}: one layer architecture with direct prediction of object height and camera height without refinements. (2) \emph{SN-L3}: one layer for initial prediction with 2 additional refinement layers. (3) \emph{SN-L3-mulCat}: same as SN-L3 but with objects of multiple categories as input and trained on COCO-Scale-mulCat. (4) \emph{SN-L3-kps-can}: same as SN-L3 but training keypoints prediction and predicting each person in upright height instead of actual height. An upright ratio computed as $l_{\textrm{actual}} / l_{\textrm{upright}}$ in \fig{ratio_illu} is an approximation of the actual ration in 3D that takes into account the person's pose. It is multiplied to the predicted upright height to obtain actual height, and the height prior is applied to the predicted upright height.

\subsubsection{Training results on COCO-Scale}

We calculate the bounding reprojection error from \eqn{vt} on the test split which is an indication of 2D bounding box fits. The results are shown in \tab{vt_error} and \fig{thres_ratio}.

\begin{table}[!!t]
	\begin{minipage}[b]{0.5\linewidth}
	    \tiny
        \caption{$\mathcal{L}_{v_t}$ on COCO-Scale-person}
        \label{tab:vt_error}
        \begin{tabular}{c|c|c|ccccc}
        \toprule
        & & \tiny{PGM-fixedH} & \tiny{SN-L1} & \tiny{SN-L3} & \tiny{SN-L3-mulCat} & \tiny{SN-L3-kps-can} \\
        \midrule
        \multirow{3}{*}{$\mathcal{L}_{v_t}$} & mean & 0.1727 & 0.1502 & 0.0717 & 0.0613 & \textbf{0.0540} \\
        & std. & 0.3598 & 0.2579 & 0.1875 & 0.1874 & 0.1677\\
        & med. & 0.0793 & 0.0693 & 0.0116 & \textbf{0.0074} & 0.0094\\
        \bottomrule
        \end{tabular}

\end{minipage}\hfill
\begin{minipage}{0.4\linewidth}

    \centering
    \includegraphics[width = 1\linewidth]{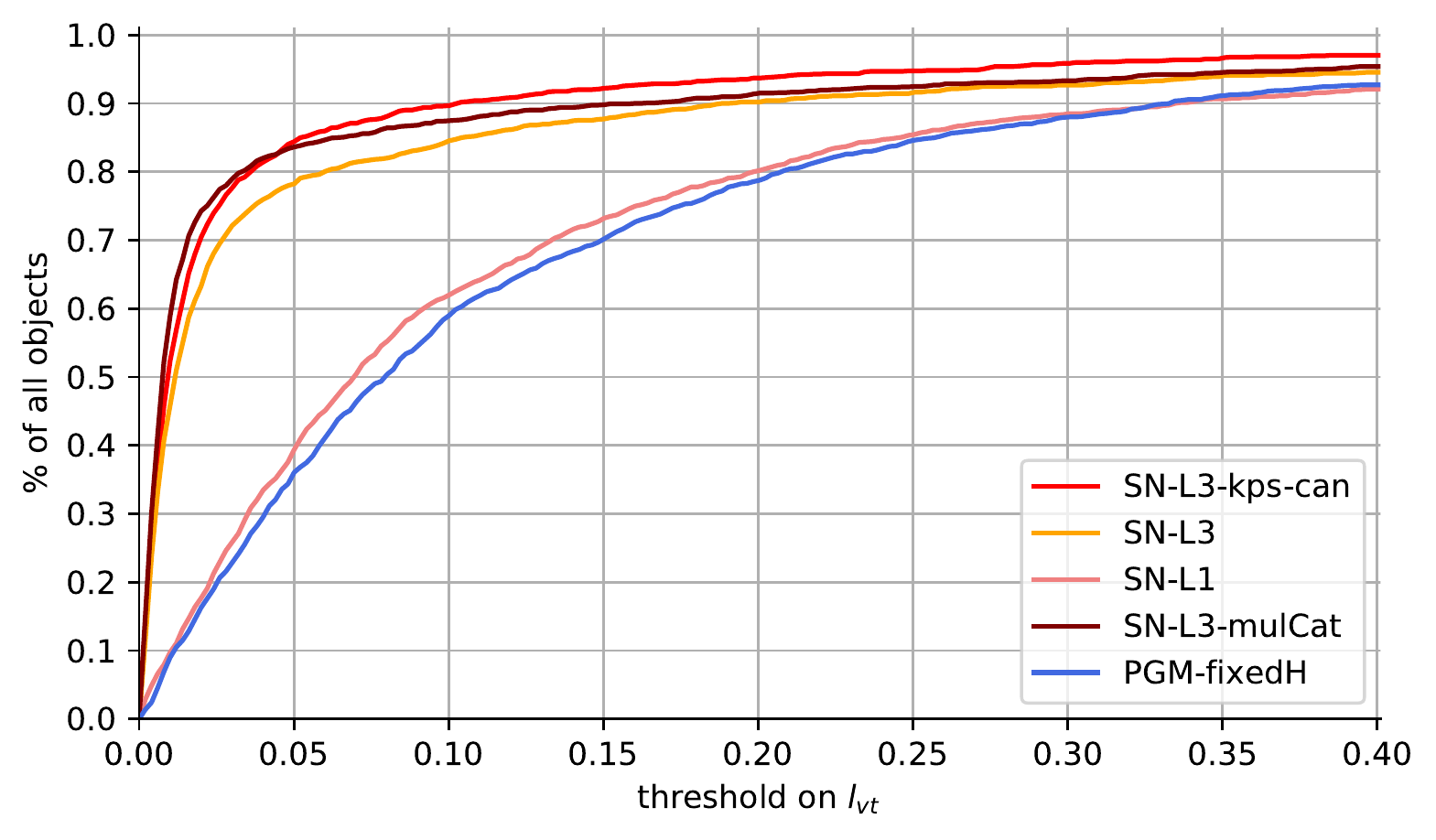}
    \captionof{figure}{$\mathcal{L}_{v_t}$ of all objects on COCO-Scale-person under varying thresholds.}
    \label{fig:thres_ratio}

\end{minipage}
\end{table}

\begin{figure*}[!!t]
    \begin{center}

\scriptsize
\begin{tabular*}{0.9\textwidth}{@{\hskip 0.15in}c |@{\hskip 0.08in}cc@{}m{0pt}@{}}
PGM & SN-L1 & SN-L3-kps-can \\ 

\image{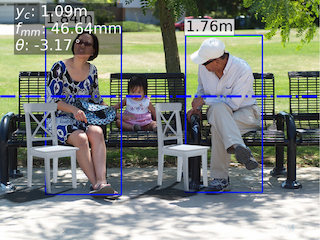} &
\image{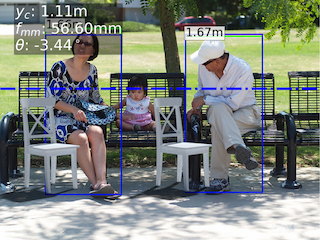} &
\image{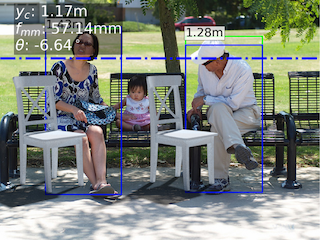} \\ 

\image{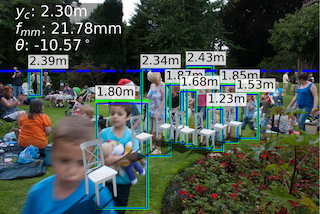} &
\image{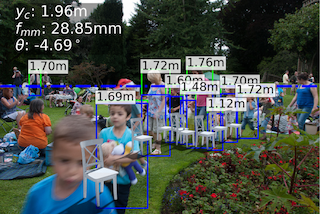} &
\image{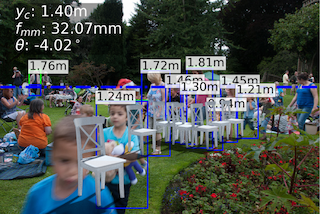} \\


\image{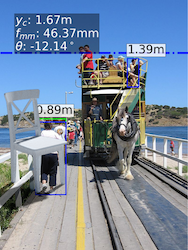} &
\image{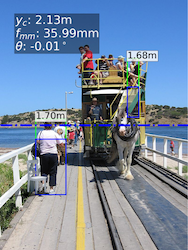} &
\image{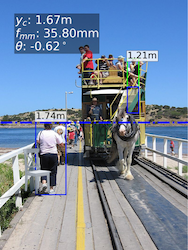} \\ 






\image{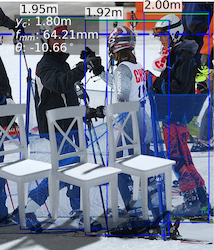} &
\image{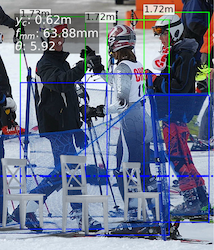} &
\image{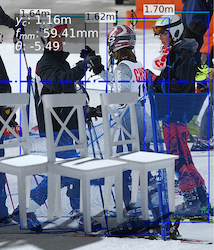} \\

\end{tabular*}

\end{center}
\vspace{-0.7cm}
\caption{Scene parameters estimation and virtual object insertion results on COCO-Scale. The detected boxes are shown in green and reprojected ones in blue. The horizon is shown as dashed blue line. Camera parameters are overlaid on the top (camera height as $y_c$, focal length in millimeters as $f_\textrm{mm}$ assuming 35mm full-frame sensor, and pitch as $\theta$). A chair of 1m tall is inserted alongside each person with the estimated parameters.}
\label{fig:more_coco}
\end{figure*}






\subsection{Evaluation on COCO-Scale-person}





Since we do not have ground truth 3D annotations on the \texttt{COCO-Scale-person} dataset, we evaluate performance using a user study on virtual object insertion. 
We evaluate the plausibility of our estimates on the resulting scale and perspective effects of an inserted object via an A/B test on \texttt{COCO-Scale-person}. In our evaluation, we render a 1m height chair alongside each object. For each of the 4 pairs of models, we insert the chair in 50 random test images. 10 users for each pair of models were asked to choose which image of the pair is more realistic \wrt the scales of all chairs. 

As can be seen in \tab{coco_ab}, our results improve when adding multiple categories (\emph{SN-L3-mulCat}) or regress keypoint and account for the pose while computing person height (\emph{SN-L3-kps-can}). The best variant \emph{SN-L3-kps-can} outperforms other methods significantly. This is consistent with the bounding box reprojection error in \tab{vt_error} where \emph{SN-L3-kps-can} is the best performing method.

\fig{more_coco} shows a qualitative evaluation of our method. The first row shows the benefit of using the upright ratio, leading to better estimations in \emph{SN-L3-kps-can}. The second row shows our method displays better behavior in cases of multiple objects with diverse heights. In the third row, we demonstrate robustness to outliers (the person on the bus). 

\setlength{\tabcolsep}{4pt}
\begin{table}[!!t]

\begin{center}
\caption{A/B test results for scale on COCO-Scale-person}
\label{tab:coco_ab}
\scriptsize
\begin{tabular}{c|cc|cc|cc|cc}
\toprule
& \tiny{SN-L3} & \tiny{SN-L1} & \tiny{SN-L3} & \tiny{SN-L3-kps-can} & \tiny{SN-L3} & \tiny{SN-L3-mulCat} & \tiny{SN-L3-kps-can} & \tiny{PGM}\\

\midrule
\multirow{1}{*}{preference} & \textbf{54.6\%} & 43.8\% & 42.8\% & \textbf{57.2\%} & \textbf{50.7}\% & 49.3\% & \textbf{59.5\%} & 40.5\%\\

\bottomrule
\end{tabular}
\end{center}
\end{table}
\setlength{\tabcolsep}{1.4pt}


\setlength{\tabcolsep}{4pt}
\begin{table}[!!t]
\begin{center}
\caption{Evaluation errors on all \textit{pedestrian} objects of KITTI-person. PGM-fixedH assumes a canonical object height,
while PGM explicitly solves for the object height that minimizes bounding box reprojection error $\mathcal{L}_{v_t}$ (ideally to zero), as a result of which the $\mathcal{L}_{v_t}$ errors of PGM are grayed out as they are not directly comparable to others.
Best results in bold (lower is better).}
\label{tab:errors_kitti}
\scriptsize
\begin{tabular}{c|c|cc|cccc}
\toprule
& & \tiny{PGM-fixedH} & \tiny{PGM} & \tiny{SN-L1} & \tiny{SN-L3} & \tiny{SN-L3-kps-can} & \tiny{SN-L3-mulCat}\\
\midrule
\multirow{2}{*}{input} & car & & & & & & \\
 & person & \checkmark & \checkmark & \checkmark & \checkmark & \checkmark & \checkmark\\
\midrule
\multirow{3}{*}{$\mathcal{E}_{h_{\textrm{obj}}}$} & mean & 0.0863 & 0.1358 & \textbf{0.0837} & 0.0956 & 0.1014 & 0.0849 \\
& std. & {0.0570} & 0.1406 & 0.0610 & 0.0751 & 0.0864 & 0.0714 \\
& med. & {0.0800} & 0.0916 & 0.0727 & 0.0770 & 0.0864 & \textbf{0.0685} \\
\midrule
\multirow{3}{*}{$\mathcal{L}_{v_t}$} & mean & 0.0767 & \gray{0.0016} & 0.0980 & 0.0331 & 0.0585 & \textbf{0.0283} \\
& std. & 0.0638 & \gray{0.0009} & 0.1000 & 0.0644 & 0.0815 & 0.0415 \\
& med. & 0.0618 & \gray{0.0003} & 0.0724 & 0.0128 & 0.0815 & \textbf{0.0127} \\
\midrule
\multirow{3}{*}{$\mathcal{E}_{h_{\textrm{cam}}}$} & mean & \multicolumn{2}{c|}{0.1408} & 0.2356 & 0.1988 & 0.2649 & \textbf{0.1264} \\
& std. & \multicolumn{2}{c|}{0.1585} & 0.2860 & 0.3162 & 0.3207 & 0.1147 \\
& med. & \multicolumn{2}{c|}{0.1096} & 0.1821 & 0.1160 & 0.1666 & \textbf{0.0878} \\
\bottomrule
\end{tabular}
\end{center}
\end{table}
\setlength{\tabcolsep}{1.4pt}

\subsection{Quantitative evaluation on KITTI}

Since KITTI provides ground truth for all of the parameters our method estimate, we can directly evaluate the errors in bounding box reprojection $\mathcal{L}_{v_t}$, camera height estimation $\mathcal{E}_{h_{\textrm{cam}}}$ and object height estimation $\mathcal{E}_{h_{\textrm{obj}}}$ as shown in \tab{errors_kitti} and \tab{errors_kitti_multi} on KITTI-person and KITTI-mulCat respectively, where 

\begin{align} \label{eqn:metrics}
\mathcal{E}_{h_{\textrm{cam}}} = \norm{h_{\textrm{cam}} - h_{\textrm{cam}_\textrm{gt}}}, \mathcal{E}_{h_{\textrm{obj}}} = \frac{1}{N}\sum_{i=1}^N\norm{h_{\textrm{obj}} - h_{\textrm{obj}_\textrm{gt}}}
\end{align}
$h_{\textrm{cam}}$ and $h_{\textrm{obj}}^i$ are the final estimated camera height and height of object $i$, and $h_{\textrm{cam}_\textrm{gt}}$ and $h_{\textrm{obj}_\textrm{gt}}^i$ are their ground truth values respectively.

\begin{figure*}[t!!]
    \begin{center}
    \setlength{\tabcolsep}{0.05em}

\small
{\renewcommand{\arraystretch}{1.2}
\begin{tabular*}{1\textwidth}[t]{cc@{}m{0pt}@{}}
 
\imageT{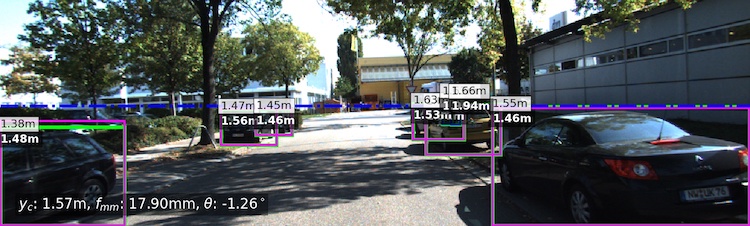} &
\imageT{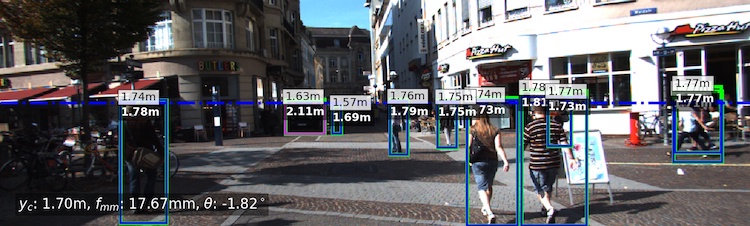} \\ 
\imageT{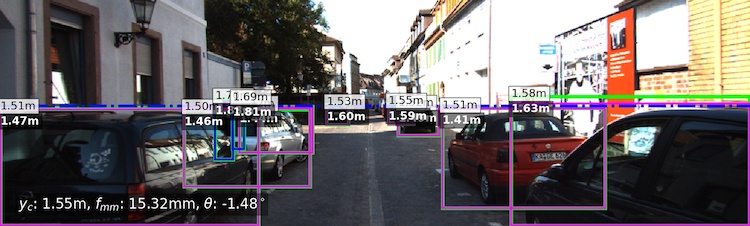} &
\imageT{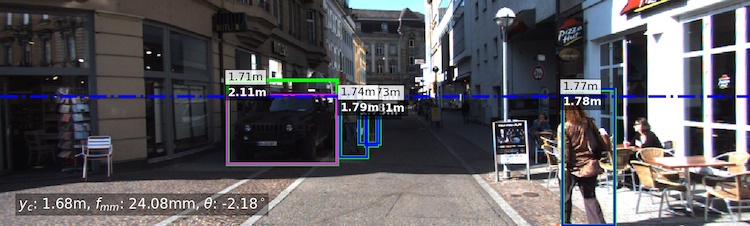} \\ 
\imageT{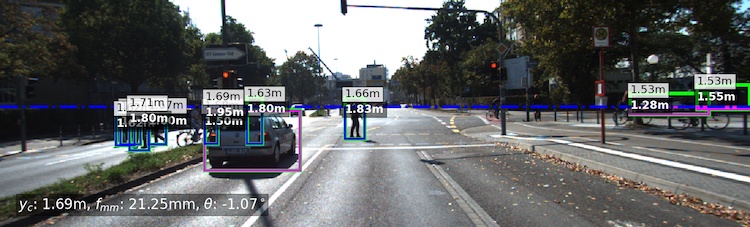} &
\imageT{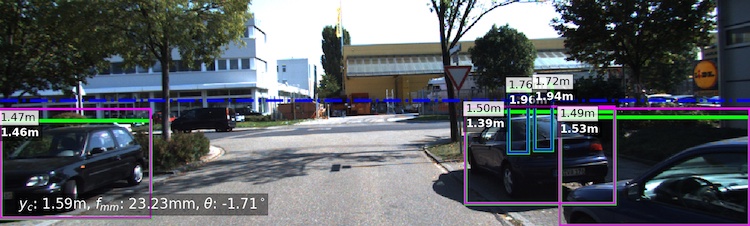} \\











\end{tabular*}
}
\end{center}
\vspace{-0.7cm}
\caption{Scene parameters estimation results with SN-L3-mulCat on KITTI-mulCat. Reprojected pedestrians are in blue, while cars are in magenta.}
\label{fig:more_KITTI}
\vspace{-0.3cm}
\end{figure*}

Our method SN-L3-mulCat outperforms previous work on both KITTI-person and KITTI-mulCat. This method takes into account the cues from multiple categories to perform inference, giving it an advantage on scenes with high-diversity content (see \tab{errors_kitti_multi}). 
Qualitative results are shown in \fig{more_KITTI}. Please refer to the supplementary material for more results.


\vspace{-0.3cm}


\setlength{\tabcolsep}{4pt}
\begin{table}[!!t]
\begin{center}
\caption{Evaluation errors on all \textit{pedestrian} objects of KITTI-mulCat following specifications of Table 3}
\label{tab:errors_kitti_multi}
\scriptsize
\begin{tabular}{c|c|cc|ccc}
\toprule
& & \tiny{PGM} & \tiny{PGM} & \tiny{SN-L3} & \tiny{SN-L3-mulCat} & \tiny{SN-L3-mulCat}\\
\midrule
\multirow{2}{*}{input} & car & & \checkmark & &  & \checkmark \\
 & person & \checkmark & \checkmark & \checkmark & \checkmark & \checkmark \\
\midrule
\multirow{3}{*}{$\mathcal{E}_{h_{\textrm{obj}}}$} & mean & 0.1198 & 0.1177 & 0.1266 & 0.1092 & \textbf{0.0956}  \\
& std. & 0.1007 & 0.0932 & 0.0994 & 0.0883 & 0.0811 \\
& med. & 0.0896 & 0.0939 & 0.0968 & 0.0876 & \textbf{0.0780} \\
\midrule
\multirow{3}{*}{$\mathcal{L}_{v_t}$} & mean & \gray{0.0008} & \gray{0.0008} & 0.0647 & \textbf{0.0303}& 0.0712 \\
& std. & \gray{0.0013} & \gray{0.0011} & 0.1124 & 0.0465 & 0.1153  \\
& med. & \gray{0.0003} & \gray{0.0004} & 0.0166  & \textbf{0.0123} & 0.0297\\
\midrule
\multirow{3}{*}{$\mathcal{E}_{h_{\textrm{cam}}}$} & mean & 0.1379 & 0.1519 & 0.3464  & 0.1547 & \textbf{0.1222}\\
& std. & 0.1735 & 0.1676 & 0.3693 & 0.1687 & 0.1235 \\
& med. & \textbf{0.0703} & 0.1096 & 0.2278 & 0.0991 & 0.0904 \\
\bottomrule
\end{tabular}
\end{center}
\end{table}
\setlength{\tabcolsep}{1.4pt}



\subsection{Quantitative evaluation on IMDB-23K}

IMDB-23K provides annotation for the registered height of a person, which we assume is the height of the person standing straight (the upright height). Since all of our models except SN-L3-kps-can predict the actual person height (influenced by the specific pose, viewpoint, bounding box drawing, etc.), we use the upright ratio (see \fig{ratio_illu}) computed from detected keypoints to convert the actual height back to upright height. This upright ratio allows us to compute upright height from all methods, and compare against the pseudo ground truth, as included in \tab{h_error_IMDB} and \fig{more_IMDB}. Since the ground truth annotations are only valid for standing people, we further get a subset of the test set where the estimated upright ratio from keypoint prediction is less than 0.90, which typically denotes a non-standing person. \tab{h_error_IMDB_sitting} evaluates the methods on this subset and shows that SN-L3-kps-can, which directly accounts for the upright ratio in training and inference, performs better in getting upright heights compared to other models; visual comparisons are shown in Fig.~\ref{fig:more_IMDB} and the supplementary material. 

\setlength{\tabcolsep}{4pt}
\begin{table}[!!t]

\begin{center}
\caption{$\mathcal{E}_{h_{\textrm{obj}}}$ and $\mathcal{L}_{v_t}$ on IMDB-23K following specifications of Table 3}
\label{tab:h_error_IMDB}
\scriptsize
\begin{tabular}{c|c|cc|cccc}
\toprule
& & \tiny{PGM-fixedH} & \tiny{PGM} & \tiny{SN-L1} & \tiny{SN-L3} & \tiny{SN-L3-kps-can} & \tiny{SN-L3-mulCat}\\
\midrule
\multirow{3}{*}{$\mathcal{E}_{h_{\textrm{obj}}}$} & mean & {0.0843} & 0.2234 & \textbf{0.0832} & 0.0891 & 0.1003 & 0.0990 \\
& std. & {0.0638} & 0.2246 & 0.0688 & 0.0818 & 0.0920 & 0.0915 \\
& med. & {0.0700} & 0.1644 & 0.0706 & \textbf{0.0695} & 0.0920 & 0.0777 \\
\midrule
\multirow{3}{*}{$\mathcal{L}_{v_t}$} & mean & 0.0983 & \gray{0.0157} & 0.1011 & 0.0431 & 0.0920 & \textbf{0.0416}\\
& std. & 0.0546 & \gray{0.0185} & 0.1706 & 0.1324 & 0.1257 & 0.1537\\
& med. & 0.0689 & \gray{0.0105} & 0.0441 & 0.0071 & 0.1257 & \textbf{0.0056}\\
\bottomrule
\end{tabular}
\end{center}
\end{table}
\setlength{\tabcolsep}{1.4pt}



\setlength{\tabcolsep}{4pt}
\begin{table}[!!t]

\begin{center}
\caption{$\mathcal{E}_{h_{\textrm{obj}}}$ on IMDB-23K (non-standing person).Best results in bold (lower is better)}
\label{tab:h_error_IMDB_sitting}
\scriptsize
\begin{tabular}{c|c|ccc}
\toprule
& & \tiny{PGM} & \tiny{SN-L3} & \tiny{SN-L3-kps-can} \\
\midrule
\multirow{3}{*}{$\mathcal{E}_{h_{\textrm{obj}}}$} & mean & 0.1552 & 0.1591 & \textbf{0.1212} \\
& std. & 0.1379 & 0.1788 & 0.1072 \\
& med. & 0.1177 & 0.1013 & \textbf{0.0909} \\
\bottomrule
\end{tabular}
\end{center}
\end{table}
\setlength{\tabcolsep}{1.4pt}

\begin{figure*}[!!t]
    \begin{center}
    \setlength{\tabcolsep}{0.05em}

\small
{\renewcommand{\arraystretch}{1.2}
\begin{tabular*}{1.\textwidth}[t]{c ccc@{}m{0pt}@{}}
 
\imageF{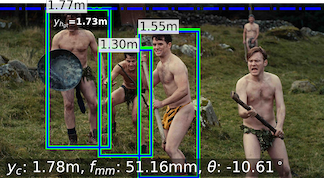} &
\imageF{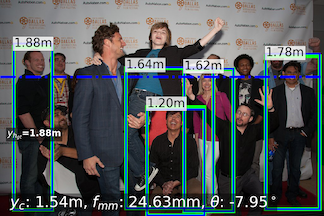} &
\imageF{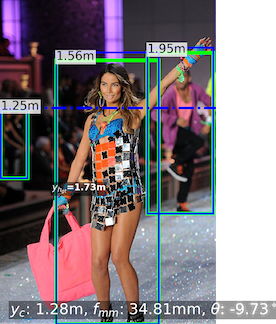} & 
\imageF{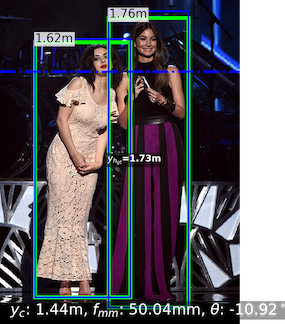}\\











\end{tabular*}
}
\end{center}
\vspace{-0.7cm}
\caption{Scene parameters estimation results with SN-L3-kps-can on IMDB-23K.}
\label{fig:more_IMDB}
\end{figure*}






\section{Conclusion and Future Work}

We present a learning-based method that performs geometric camera calibration with absolute scale from images in the wild. We demonstrate that our method provides state-of-the-art results on multiple datasets. Despite this advance, our method is hindered by some limitations. 

Our single dominant ground plane assumption does not always hold in the wild. Urban scene may provide multiple supporting surfaces at different heights (tables, balconies), so objects may not be laying on the assumed ground plane. Also, the ground may be non-flat in nature environments.

Our method is highly  biased on \emph{appearance}. Adding amodal reasoning, as proposed in~\cite{kar2015amodal}, would be an interesting way forward to perform holistic scene reasoning. We would like to tackle these limitations as future work.


\clearpage
%
%
\bibliographystyle{splncs04}
\bibliography{egbib}
\end{document}